\documentclass[11pt,a4paper]{article}
\usepackage[hyperref]{acl2019}
\usepackage{times}
\usepackage{footnote}
\usepackage{latexsym}
\usepackage{xcolor}
\usepackage{url}
\usepackage{graphicx} 
\usepackage{booktabs}
\usepackage{multirow}
\usepackage{amsmath}
\usepackage{color, soul}
\usepackage{amssymb}
\usepackage{pifont}
\usepackage{cleveref}[2012/02/15]
\newcommand{\xmark}{\ding{55}}

\newcommand{\hlc}[2][yellow]{ {\sethlcolor{#1} \hl{#2}} }
\newcommand{\greenv}{\textcolor{green}{\checkmark}} 
\newcommand{\redx}{\textcolor{red}{\xmark}}

\crefformat{footnote}{#2\footnotemark[#1]#3}

\aclfinalcopy 

\title{\textsc{TalkSumm}: A Dataset and Scalable Annotation Method for Scientific Paper Summarization Based on Conference Talks}

\author{Guy Lev\thanks{\;\;\; The authors contributed equally.}$^{*}$, Michal Shmueli-Scheuer$^{*}$, Jonathan Herzig, Achiya Jerbi, David Konopnicki\\
  IBM Research, Haifa, Israel \\
  \{guylev,shmueli,hjon,davidko\}@il.ibm.com, achiya.jerbi@ibm.com
  }
\date{}

\begin{document}
\maketitle
\begin{abstract}
Currently, no large-scale training data is available for the task of scientific paper summarization. 
In this paper, we propose a novel method that automatically generates summaries for scientific papers, by utilizing videos of talks at scientific conferences. We hypothesize that such talks constitute a coherent and concise description of the papers' content, and can form the basis for good summaries. 
We collected $1716$ papers and their corresponding videos, and created a dataset of paper summaries. A model trained on this dataset achieves similar performance as models trained on a dataset of summaries created manually. In addition, we validated the quality of our summaries by human experts.
\end{abstract}

\section{Introduction}
\label{intro}

The rate of publications of scientific papers is increasing and it is almost impossible for researchers to keep up  with relevant research. Automatic text summarization could help mitigate this problem.
In general, there are two common approaches to summarizing scientific papers: \textit{citations-based}, based on a set of citation sentences~\cite{Nakov04citances,Abu-Jbara:2011,aaai19.scisumm}, and \textit{content-based}, 
based on the paper itself~\cite{CoNLLabstractive,NIKOLOV18.2}. Automatic summarization is studied exhaustively for the news domain~\cite{Lapata2016,see2017}, 
while summarization of scientific papers is less studied, mainly due to the lack of large-scale training data. 
The papers' length and complexity require substantial summarization effort from experts.
Several methods were suggested to reduce these efforts~\cite{aaai19.scisumm,CoNLLabstractive}, still they are not scalable as they require human annotations.

\definecolor{dandelion}{rgb}{0.94, 0.88, 0.19}
\definecolor{green-yellow}{rgb}{0.68, 1.0, 0.18}
\definecolor{mediumspringbud}{rgb}{0.79, 0.86, 0.54}
\definecolor{palecerulean}{rgb}{0.61, 0.77, 0.89}
\definecolor{paleplum}{rgb}{0.8, 0.6, 0.8}
\definecolor{carrotorange}{rgb}{0.93, 0.57, 0.13}

\begin{table}[t]
\resizebox{1.0\columnwidth}{!}{
\scriptsize
\begin{tabular}{|p{10cm}|}
\hline
\textbf{Title:} Split and Rephrase: Better Evaluation and Stronger Baselines \cite{aharoni2018split} \\
 \hline
\textbf{Paper:} \hlc[dandelion]{Processing long, complex sentences is challenging. This is true either for humans in various circumstances or in NLP tasks like parsing and machine translation}.\hlc[mediumspringbud]{An automatic system capable of breaking a complex sentence into several simple sentences that convey the same meaning is very appealing}. \hlc[palecerulean]{A recent work by Narayan et al. (2017) introduced a dataset, evaluation method and baseline systems for the task, naming it Split-and Rephrase}. The dataset includes 1,066,115 instances mapping a single complex sentence to a sequence of sentences that express the same meaning, together with RDF triples that describe their semantics. They considered two \ldots Indeed, feeding the model with examples containing entities alone without any facts about them causes it to output perfectly phrased but unsupported facts (Table 3). Digging further, we find that 99\% of the simple sentences (more than 89\% of the unique ones) in the validation and test sets \hlc[paleplum]{also appear in the training set, which coupled with the good memorization capabilities of SEQ2SEQ models and the relatively small number of distinct simple sentences helps to explain the high BLEU score}. \hlc[carrotorange]{To aid further research on the task, we propose a more challenging split of the data}. We also establish a stronger baseline by extending the SEQ2SEQ approach with a copy mechanism, which was shown \ldots
 We encourage future work on the split-and-rephrase task to use our new data split or the v1.0 split instead of the original one. \\
\hline
 \textbf{Talk transcript:} let's begin with the motivation so \hlc[dandelion]{processing long complex sentences is a hard task this is true for arguments like children people with reading disabilities second language learners but this is also true for sentence level and NLP systems}, for example previous work show that \hlc[dandelion]{dependency parsers} degrade performance when they're introduced with longer and longer sentences, in a \hlc[dandelion]{similar result was shown for neural machine translation}, where neural machine translation systems introduced with longer sentences starting degrading performance, \hlc[mediumspringbud]{the question rising here is can we automatically break a complex sentence into several simple ones while preserving the meaning or the semantics and this can be a useful component in NLP pipelines}. For example, \hlc[palecerulean]{the split and rephrase task was introduced in the last EMNLP by Narayan, Gardent and Shimarina, where they introduced a dataset, an evaluation method and baseline models for this task. The task definition can be taking a complex sentence and breaking it into several simple ones with the same meaning}. For example, \ldots
 semantics units in the source sentence and then rephrasing those units into a single sentences on the target site. In this work we first show \hlc[paleplum]{the simple neural models seem to perform very well on the original benchmark, but this is only due to memorization of the training set}, \hlc[carrotorange]{we propose a more challenging data split for the task} to discourage this memorization and we perform automatic evaluation in error analysis on the new benchmark showing that the task is still very far from being solved. \\ 
 \hline
\end{tabular}}
  \caption{Alignment example between a paper's Introduction section and first 2:40 minutes of the talk's transcript. The different colors show corresponding content between the transcript to the written paper.}~\label{fig:example}
\end{table}

Recently, academic conferences started publishing videos of talks (e.g., ACL\footnote{\label{note1}\url{vimeo.com/aclweb}}, EMNLP\cref{note1}, ICML\footnote{\url{icml.cc/Conferences/2017/Videos}}, and more).
In such talks, the presenter (usually a co-author) must describe their paper coherently and concisely (since there is a time limit), providing a good basis for generating summaries. 
Based on this idea, in this paper, we propose a new method, named \textsc{TalkSumm} (acronym for \textit{Talk-based Summarization}), to automatically generate extractive content-based summaries for scientific papers based on video talks.
Our approach utilizes the transcripts of video content of conference talks, and treat them as spoken summaries of papers. Then, using unsupervised alignment algorithms, we map the transcripts to the corresponding papers' text, and create extractive summaries. 
Table~\ref{fig:example} gives an example of an alignment between a paper and its talk transcript (see Table~\ref{fig:example_gt_alignment} in the appendix for a complete example).

Summaries generated with our approach can then be used to train more complex and data-demanding summarization models. 
Although our summaries may be noisy (as they are created automatically from transcripts), our dataset can easily grow in size as more
 conference videos are aggregated. Moreover, our approach can generate summaries of various lengths.

Our main contributions are as follows: (1) we propose a new approach to automatically generate summaries for scientific papers based on video talks; (2) we create a new dataset, that contains $1716$ summaries for papers from several computer science conferences, that can be used as training data; (3) we show both automatic and human evaluations for our approach. We make our dataset and related code publicly available\footnote{\url{https://github.com/levguy/talksumm}}.
To our knowledge, this is the first approach to automatically create extractive summaries for scientific papers by utilizing the videos of conference talks. 

\section{Related Work}
\label{related}

Several works focused on generating training data for scientific paper summarization~\cite{aaai19.scisumm,scisumm18,CoNLLabstractive,Cohan2018}.
Most prominently, the CL-SciSumm shared tasks~\cite{scisumm16,scisumm18} provide a total of 40 human generated summaries;
there, a citations-based approach is used, where experts first read citation sentences (citances) that reference the paper being summarized, and then read the whole paper. Then, they create a summary of 150 words on average. 

Recently, to mitigate annotation cost, \newcite{aaai19.scisumm} proposed a  
method, in which human annotators only read the abstract in addition to citances (not reading the full paper). Using this approach, they generated $1000$ summaries, costing 600+ person-hours. Conversely, we generate summaries, given transcripts of conference talks, in a fully automatic manner, and, thus, our approach is much more scalable. ~\newcite{CoNLLabstractive} also aimed at generating labeled data for scientific paper summarization, 
based on ``highlight statements'' that authors can provide in some publication venues. 

Using external data to create summaries was also proposed in the news domain. ~\newcite{microblogSum, tweetsSumSigir2015} utilized tweets to decide which sentences to extract from news article.

Finally, alignment between different modalities (e.g., presentation, videos) and text was studied in different domains. Both \newcite{Kan:2007} and \newcite{Bahrani:2013} studied the problem of document to presentation alignment for scholarly documents.  \newcite{Kan:2007} focused on the the discovery and crawling of document-presentation pairs, and a model to align between documents to corresponding presentations. In \newcite{Bahrani:2013} they extended previous model to include also visual components of the slides. 
Aligning video and text was studied mainly in the setting of enriching 
videos with textual information~\cite{Bojanowski_2015_ICCV,googleCooking,Zhu_2015_ICCV}. 
 ~\newcite{googleCooking} used HMM to align ASR transcripts of cooking videos and recipes text for enriching videos with instructions. \newcite{Zhu_2015_ICCV} utilized books to enrich videos with descriptive explanations.
\newcite{Bojanowski_2015_ICCV} proposed to align video and text by providing a time stamp for every sentence. The main difference between these works and ours is in the alignment being used to generate textual training data in our case, rather than to enrich videos.

\section{The \textsc{TalkSumm} Dataset}
\label{data}

\subsection{Data Collection}

Recently, many computer science academic associations including ACL, ACM, IMLS and more, have started recording talks in different conferences, e.g., ACL, NAACL, EMNLP, and other co-located workshops.
A similar trend occurs in other domains such as Physics\footnote{\url{www.cleoconference.org}}, Biology\footnote{\url{igem.org/Videos/Lecture_Videos}}, etc.

In a conference, each speaker (usually a co-author) presents their paper given a timeframe of 15-20 minutes.
Thus, the talk must be coherent and concentrate on the most important aspects of a paper.
Hence, the talk can be considered as a summary of the paper, as viewed by its authors, 
and is much more comprehensive
than the abstract, which is written by the authors as well.

In this work, we focused on NLP and ML conferences, and analyzed $1716$ video talks from ACL, NAACL, EMNLP, SIGDIAL (2015-2018), and ICML (2017-2018).
We downloaded the videos and extracted the speech data.
Then, via a publicly available ASR service\footnote{\url{www.ibm.com/watson/services/speech-to-text/}}, we extracted transcripts of the speech, and based on the video metadata (e.g., title), we retrieved the corresponding 
paper (in PDF format). We used Science-Parse\footnote{\url{github.com/allenai/science-parse}} to extract the text of the paper, and applied a simple processing in order to filter-out some noise (e.g. lines starting with the word ``Copyright'').
At the end of this process, the text of each paper is associated with the transcript of the corresponding talk.

\subsection{Dataset Generation}

The transcript itself cannot serve as a good summary for the corresponding paper, as it constitutes only one modality of the talk (which also consists of slides, for example), and hence cannot stand by itself and form a coherent written text.
Thus, to create an extractive paper summary based on the transcript, we model the alignment between spoken words and sentences in the paper, assuming the following generative process: 
During the talk, the speaker generates words for describing verbally sentences from the paper, one word at each time step. Thus, at each time step, the speaker has a single sentence from the paper in mind, and produces a word that constitutes a part of its verbal description.
Then, at the next time-step, the speaker either stays with the same sentence, or moves on to describing another sentence, and so on.
Thus, given the transcript, we aim to retrieve those ``source'' sentences and use them as the summary. The number of words uttered to describe each sentence can serve as importance score, indicating the amount of time the speaker spent describing the sentence. This enables to control the summary length by considering only the most important sentences up to some threshold.

We use an HMM to model the assumed generative process. The sequence of spoken words is the output sequence. Each hidden state of the HMM corresponds to a single paper sentence. We heuristically define the HMM's probabilities as follows.

Denote by $Y(1:T)$ the spoken words, and by $S(t)\in\{1,...,K\}$ the paper sentence index at time-step $t\in\{1,...,T\}$.
Similarly to \newcite{googleCooking}, we define the emission probabilities to be:
\begin{align*}
p&(Y(t)=y|S(t)=k) \propto \nonumber & \max_{w\in words(k)}sim(y,w)
\end{align*}

where $words(k)$ is the set of words in the $k$'th sentence, and $sim$ is a semantic-similarity measure between words, based on word-vector distance. We use pre-trained GloVe \cite{Pennington14glove:global} as the semantic vector representations for words.

As for the transition probabilities, we must model the speaker's behavior and the transitions between any two sentences in the paper. This is unlike the simpler setting in \newcite{googleCooking}, where transition is allowed between consecutive sentences only. To do so, denote the entries of the transition matrix by $T(k,l)=p(S(t+1)=l|S(t)=k)$.
We rely on the following assumptions: (1) $T(k,k)$ (the probability of staying in the same sentence at the next time-step) is relatively high. (2) There is an inverse relation between $T(k,l)$ and $|l-k|$, i.e., it is more probable to move to a nearby sentence than jumping to a farther sentence.
(3)  $S(t+1)> S(t)$ is more probable than the opposite (i.e., transition to a later sentence is more probable than to an earlier one).
Although these assumptions do not perfectly reflect reality, they are a reasonable approximation in practice.

Following these assumptions, we define the HMM's transition probability matrix. First, define the \textit{stay-probability} as $\alpha=\max(\delta(1-\frac{K}{T}), \epsilon)$, where $\delta,\epsilon\in(0,1)$. This choice of stay-probability is inspired by \newcite{googleCooking}, using $\delta$ to fit it to our case where transitions between any two sentences are allowed, and $\epsilon$ to handle rare cases where $K$ is close to, or even larger than $T$. Then, for each sentence index $k\in\{1,...,K\}$, we define:

\[T(k,k)=\alpha\]
\[T(k,k+j)=\beta_k\cdot\lambda^{j-1}, \quad j\geq 1\]
\[T(k,k-j)=\gamma\cdot\beta_k\cdot\lambda^{j-1}, \quad j\geq 1\]
where $\lambda,\gamma,\beta_k\in(0,1)$, $\lambda$ and $\gamma$ are factors reflecting assumptions (2) and (3) respectively, and for all $k$, $\beta_k$ is normalized s.t. $\sum_{l=1}^{K} T(k,l) = 1$. The values of $\lambda$, $\gamma$, $\delta$ and $\epsilon$ were fixed throughout our experiments at $\lambda=0.75$, $\gamma=0.5$, $\delta=0.33$ and $\epsilon=0.1$. The average value of $\alpha$, across all papers, was around $0.3$. The values of these parameters were determined based on evaluation over manually-labeled alignments between the transcripts and the sentences of a small set of papers.

Finally, we define the start-probabilities assuming that the first spoken word must be conditioned on a sentence from the Introduction section, hence $p(S(1))$ is defined as a uniform distribution over the Introduction section's sentences.

Note that sentences which appear in the Abstract, Related Work, and Acknowledgments sections of each paper are excluded from the HMM's hidden states, as we observed that presenters seldom refer to them.

To estimate the MAP sequence of sentences, we apply the Viterbi algorithm.
The sentences in the obtained sequence are the candidates for the paper's summary. For each sentence $s$ appearing in this sequence, denote by $count(s)$ the number of time-steps in which this sentence appears. Thus, $count(s)$ models the number of words generated by the speaker conditioned on $s$, and, hence, can be used as an importance score. Given a desired summary length, one can draw a subset of top-ranked sentences up to this length.

\section{Experiments}
\label{exp}

\subsection{Experimental Setup}
\paragraph{Data For Evaluation}
We evaluate the quality of our dataset generation method by training an extractive summarization model, and evaluating this model on a human-generated dataset of scientific paper summaries. For this, we choose 
 the CL-SciSumm shared task~\cite{scisumm16,scisumm18}, as this is the most established benchmark for scientific paper summarization.  
In this dataset, experts
wrote summaries of 150 words length on average, after reading the whole paper. The evaluation is on the same test data used by \newcite{aaai19.scisumm}, namely 
10 examples from CL-SciSumm 2016, and 20 examples from CL-SciSumm 2018 as validation data.

\paragraph{Training Data}
Using the HMM importance scores, we create four training sets, two with fixed-length summaries (150 and 250 words), and two with fixed ratio between summary and paper lengths (0.3 and 0.4). We train models on each training set, and select the model yielding the best performance on the validation set (evaluation is always done with generating a 150-words summary).

\paragraph{Summarization Model}
We train an extractive summarization model on our \textsc{TalkSumm} dataset, using the extractive variant of \newcite{fast2018chen}.
We test two summary generation approaches, similarly to \newcite{aaai19.scisumm}. First, for \textsc{TalkSumm-only}, we generate a 150-words summary out of the top-ranked sentences extracted by our trained model (sentences from the Acknowledgments section are omitted, in case the model extracts any). In the second approach, a 150-words summary is created by augmenting 
the abstract with non-redundant sentences extracted by our model, similarly to the ``Hybrid 2'' approach of \newcite{aaai19.scisumm}.
We perform early-stopping and hyper-parameters tuning using the validation set.

\paragraph{Baselines}
We compare our results to \textsc{ScisummNet} \cite{aaai19.scisumm} trained on 1000 scientific papers summarized by human annotators. 
As we use the same test set as in \newcite{aaai19.scisumm}, we directly compare their reported model performance to ours, including their \textsc{Abstract} baseline which takes the abstract to be the paper's summary.

\subsection{Results}
\paragraph{Automatic Evaluation}

Table~\ref{tab:results} summarizes the results: both \textsc{GCN Cited text spans} and \textsc{TalkSumm-only} models,  are not able to obtain better performance than \textsc{Abstract}\footnote{While the abstract was input to \textsc{GCN Cited text spans}, it was excluded from \textsc{TalkSumm-only}.}.
However, for the Hybrid approach, where the abstract is augmented with sentences from the summaries emitted by the models, our \textsc{TalkSumm-Hybrid} outperforms both \textsc{GCN Hybrid 2} and \textsc{Abstract}.
Importantly, our model, trained on automatically-generated summaries, performs on par with models trained over \textsc{ScisummNet}, in which training data was created manually.

\begin{table}[t]
\centering
\resizebox{1.0\columnwidth}{!}{
\begin{tabular}{l||l|c|c|c}
\hline\hline
Model                & 2-R   & 2-F   & 3-F   & SU4-F \\
\textsc{TalkSumm-Hybrid}             & \textbf{35.05} & \textbf{34.11} & \textbf{27.19} & \textbf{24.13} \\
\textsc{TalkSumm-only}            & 22.77 & 21.94 & 15.94 & 12.55 \\ \hline
\textsc{GCN Hybrid 2}*      &  32.44 & 30.08 & 23.43 & 23.77 \\
\textsc{GCN Cited text spans}* & 25.16 & 24.26 & 18.79 & 17.67 \\
\textsc{Abstract}*       & 29.52 & 29.4  & 23.16 & 23.34 \\ \hline\hline
\end{tabular}}
\caption{ROUGE scores on the CL-SciSumm 2016 test benchmark. *: results from \newcite{aaai19.scisumm}.}
\label{tab:results}
\end{table}

\paragraph{Human Evaluation}

We conduct a human evaluation of our approach with support from authors who presented their papers in conferences. 
As our goal is to test more comprehensive summaries, we generated summaries composed of $30$ sentences (approximately $15\%$ of a long paper).
We randomly selected $15$ presenters from our corpus and asked them to perform two tasks, given the generated summary of their paper: (1) for each sentence in the summary, we asked them to indicate whether they considered it when preparing the talk (yes/no question); (2) we asked them to globally evaluate the quality of the summary (1-5 scale, ranging from very bad to excellent, 3 means good). 
\\
For the sentence-level task (1), $73\%$ of the sentences were considered while preparing the talk. As for the global task (2), the quality of the summaries was $3.73$ on average, with standard deviation of $0.725$. These results validate the quality of our generation method. 
\section{Conclusion}
\label{conclusion}

We propose a novel automatic method to generate training data for scientific papers summarization, based on conference talks given by authors.
We show that the a model trained on our dataset achieves competitive results compared to models trained on human generated summaries, and that the dataset quality satisfies human experts.
In the future, we plan to study the effect of other video modalities on the alignment algorithm. We hope our method and dataset will unlock new opportunities for scientific paper summarization.

\bibliography{references}

\begin{thebibliography}{20}
\expandafter\ifx\csname natexlab\endcsname\relax\def\natexlab#1{#1}\fi

\bibitem[{Abu-Jbara and Radev(2011)}]{Abu-Jbara:2011}
Amjad Abu-Jbara and Dragomir Radev. 2011.
\newblock Coherent citation-based summarization of scientific papers.
\newblock In \emph{Proceedings of the 49th Annual Meeting of the Association
  for Computational Linguistics: Human Language Technologies - Volume 1}, HLT
  '11, pages 500--509.

\bibitem[{Aharoni and Goldberg(2018)}]{aharoni2018split}
Roee Aharoni and Yoav Goldberg. 2018.
\newblock \href {http://aclweb.org/anthology/P18-2114} {Split and rephrase:
  Better evaluation and stronger baselines}.
\newblock In \emph{Proceedings of the 56th Annual Meeting of the Association
  for Computational Linguistics (Volume 2: Short Papers)}, pages 719--724.
  Association for Computational Linguistics.

\bibitem[{Bahrani and Kan(2013)}]{Bahrani:2013}
Bamdad Bahrani and Min-Yen Kan. 2013.
\newblock Multimodal alignment of scholarly documents and their presentations.
\newblock In \emph{Proceedings of the 13th ACM/IEEE-CS Joint Conference on
  Digital Libraries}, JCDL '13, pages 281--284.

\bibitem[{Bojanowski et~al.(2015)Bojanowski, Lajugie, Grave, Bach, Laptev,
  Ponce, and Schmid}]{Bojanowski_2015_ICCV}
Piotr Bojanowski, Remi Lajugie, Edouard Grave, Francis Bach, Ivan Laptev, Jean
  Ponce, and Cordelia Schmid. 2015.
\newblock Weakly-supervised alignment of video with text.
\newblock In \emph{The IEEE International Conference on Computer Vision
  (ICCV)}.

\bibitem[{Chen and Bansal(2018)}]{fast2018chen}
Yen-Chun Chen and Mohit Bansal. 2018.
\newblock \href {http://aclweb.org/anthology/P18-1063} {Fast abstractive
  summarization with reinforce-selected sentence rewriting}.
\newblock In \emph{Proceedings of the 56th Annual Meeting of the Association
  for Computational Linguistics (Volume 1: Long Papers)}, pages 675--686.
  Association for Computational Linguistics.

\bibitem[{Cheng and Lapata(2016)}]{Lapata2016}
Jianpeng Cheng and Mirella Lapata. 2016.
\newblock Neural summarization by extracting sentences and words.
\newblock In \emph{Proceedings of the 54th Annual Meeting of the Association
  for Computational Linguistics (Volume 1: Long Papers)}, pages 484--494.

\bibitem[{Cohan and Goharian(2018)}]{Cohan2018}
Arman Cohan and Nazli Goharian. 2018.
\newblock Scientific document summarization via citation contextualization and
  scientific discourse.
\newblock \emph{International Journal on Digital Libraries}, pages 287--303.

\bibitem[{Collins et~al.(2017)Collins, Augenstein, and
  Riedel}]{CoNLLabstractive}
Ed~Collins, Isabelle Augenstein, and Sebastian Riedel. 2017.
\newblock A supervised approach to extractive summarisation of scientific
  papers.
\newblock In \emph{Proceedings of the 21st Conference on Computational Natural
  Language Learning (CoNLL 2017)}, pages 195--205.

\bibitem[{Jaidka et~al.(2016)Jaidka, Chandrasekaran, Rustagi, and
  Kan}]{scisumm16}
Kokil Jaidka, Muthu~Kumar Chandrasekaran, Sajal Rustagi, and Min-Yen Kan. 2016.
\newblock Overview of the cl-scisumm 2016 shared task.
\newblock In \emph{In Proceedings of Joint Workshop on Bibliometric-enhanced
  Information Retrieval and NLP for Digital Libraries (BIRNDL 2016)}.

\bibitem[{Jaidka et~al.(2018)Jaidka, Yasunaga, Chandrasekaran, Radev, and
  Kan}]{scisumm18}
Kokil Jaidka, Michihiro Yasunaga, Muthu~Kumar Chandrasekaran, Dragomir Radev,
  and Min-Yen Kan. 2018.
\newblock The cl-scisumm shared task 2018: Results and key insights.
\newblock In \emph{Proceedings of the 3rd Joint Workshop on
  Bibliometric-enhanced Information Retrieval and Natural Language Processing
  for Digital Libraries (BIRNDL)}.

\bibitem[{Kan(2007)}]{Kan:2007}
Min-Yen Kan. 2007.
\newblock Slideseer: A digital library of aligned document and presentation
  pairs.
\newblock In \emph{Proceedings of the 7th ACM/IEEE-CS Joint Conference on
  Digital Libraries}, JCDL '07, pages 81--90.

\bibitem[{Malmaud et~al.(2015)Malmaud, Huang, Rathod, Johnston, Rabinovich, and
  Murphy}]{googleCooking}
Jonathan Malmaud, Jonathan Huang, Vivek Rathod, Nicholas Johnston, Andrew
  Rabinovich, and Kevin Murphy. 2015.
\newblock What's cookin'? interpreting cooking videos using text, speech and
  vision.
\newblock In \emph{Proceedings of the 2015 Conference of the North American
  Chapter of the Association for Computational Linguistics: Human Language
  Technologies}, pages 143--152. Association for Computational Linguistics.

\bibitem[{Nakov et~al.(2004)Nakov, Schwartz, and Hearst}]{Nakov04citances}
Preslav~I. Nakov, Ariel~S. Schwartz, and Marti~A. Hearst. 2004.
\newblock Citances: Citation sentences for semantic analysis of bioscience
  text.
\newblock In \emph{In Proceedings of the SIGIR?04 workshop on Search and
  Discovery in Bioinformatics}.

\bibitem[{Nikola~Nikolov and Hahnloser(2018)}]{NIKOLOV18.2}
Michael~Pfeiffer Nikola~Nikolov and Richard Hahnloser. 2018.
\newblock Data-driven summarization of scientific articles.
\newblock In \emph{Proceedings of the Eleventh International Conference on
  Language Resources and Evaluation (LREC 2018)}.

\bibitem[{Pennington et~al.(2014)Pennington, Socher, and
  Manning}]{Pennington14glove:global}
Jeffrey Pennington, Richard Socher, and Christopher~D. Manning. 2014.
\newblock Glove: Global vectors for word representation.
\newblock In \emph{In EMNLP}.

\bibitem[{See et~al.(2017)See, Liu, and Manning}]{see2017}
Abigail See, Peter~J. Liu, and Christopher~D. Manning. 2017.
\newblock Get to the point: Summarization with pointer-generator networks.
\newblock In \emph{Proceedings of the 55th Annual Meeting of the Association
  for Computational Linguistics (Volume 1: Long Papers)}, pages 1073--1083.

\bibitem[{Wei and Gao(2014)}]{microblogSum}
Zhongyu Wei and Wei Gao. 2014.
\newblock \href {http://aclweb.org/anthology/C14-1083} {Utilizing microblogs
  for automatic news highlights extraction}.
\newblock In \emph{Proceedings of COLING 2014, the 25th International
  Conference on Computational Linguistics: Technical Papers}, pages 872--883.
  Dublin City University and Association for Computational Linguistics.

\bibitem[{Wei and Gao(2015)}]{tweetsSumSigir2015}
Zhongyu Wei and Wei Gao. 2015.
\newblock Gibberish, assistant, or master?: Using tweets linking to news for
  extractive single-document summarization.
\newblock In \emph{Proceedings of the 38th International ACM SIGIR Conference
  on Research and Development in Information Retrieval}, SIGIR '15, pages
  1003--1006.

\bibitem[{Yasunaga et~al.(2019)Yasunaga, Kasai, Zhang, Fabbri, Li, Friedman,
  and Radev}]{aaai19.scisumm}
Michihiro Yasunaga, Jungo Kasai, Rui Zhang, Alexander Fabbri, Irene Li, Dan
  Friedman, and Dragomir Radev. 2019.
\newblock Scisummnet: A large annotated corpus and content-impact models for
  scientific paper summarization with citation networks.
\newblock In \emph{Proceedings of AAAI 2019}.

\bibitem[{Zhu et~al.(2015)Zhu, Kiros, Zemel, Salakhutdinov, Urtasun, Torralba,
  and Fidler}]{Zhu_2015_ICCV}
Yukun Zhu, Ryan Kiros, Rich Zemel, Ruslan Salakhutdinov, Raquel Urtasun,
  Antonio Torralba, and Sanja Fidler. 2015.
\newblock Aligning books and movies: Towards story-like visual explanations by
  watching movies and reading books.
\newblock In \emph{The IEEE International Conference on Computer Vision
  (ICCV)}.

\end{thebibliography}
\bibliographystyle{acl_natbib}

\appendix

\section{A Detailed Example}
\label{appendix}

\definecolor{britishracinggreen}{rgb}{0.0, 0.26, 0.15}
\definecolor{darkpink}{rgb}{0.91, 0.33, 0.5}
\definecolor{bittersweet}{rgb}{1.0, 0.44, 0.37}
\definecolor{cerulean}{rgb}{0.0, 0.48, 0.65}
\definecolor{palatinatepurple}{rgb}{0.41, 0.16, 0.38}
\definecolor{olive}{rgb}{0.5, 0.5, 0.0}

\definecolor{dandelion}{rgb}{0.94, 0.88, 0.19}
\definecolor{green-yellow}{rgb}{0.68, 1.0, 0.18}
\definecolor{mediumspringbud}{rgb}{0.79, 0.86, 0.54}
\definecolor{palecerulean}{rgb}{0.61, 0.77, 0.89}
\definecolor{paleplum}{rgb}{0.8, 0.6, 0.8}
\definecolor{carrotorange}{rgb}{0.93, 0.57, 0.13}

This section elaborates on the example presented in Table~\ref{fig:example}.
Table~\ref{fig:example_gt_alignment} extends Table~\ref{fig:example} by showing the manually-labeled alignment between the complete text of the paper's Introduction section, and the corresponding transcript.
Table~\ref{fig:example_model_output} shows the alignment obtained using the HMM. Each row in this table corresponds to an interval of consecutive time-steps (i.e., a sub-sequence of the transcript) in which the same paper sentence was selected by the Viterbi algorithm. The first column (Paper Sentence) shows the selected sentences; The second column (ASR transcript) shows the transcript obtained by the ASR system; The third column (Human transcript) shows the manually corrected transcript, which is provided for readability (our model predicted the alignment based on the raw ASR output); Finally, the forth column shows whether our model has correctly aligned a paper sentence with a sub-sequence of the transcript. Rows with no values in this column correspond to transcript sub-sequences which were not associated with any paper sentence in the manually-labeled alignment.

\begin{table}[t]
\resizebox{1.0\columnwidth}{!}{
\scriptsize
\begin{tabular}{|p{10cm}|}
\hline
\textbf{Title:} Split and Rephrase: Better Evaluation and Stronger Baselines \cite{aharoni2018split} \\
 \hline
\textbf{Paper:} \hlc[dandelion]{Processing long, complex sentences is challenging. This is true either for humans in various circumstances or in NLP tasks like parsing and machine translation}. \hlc[mediumspringbud]{An automatic system capable of breaking a complex sentence into several simple sentences that convey the same meaning is very appealing}. \hlc[palecerulean]{A recent work by Narayan et al. (2017) introduced a dataset, evaluation method and baseline systems for the task, naming it Split-and Rephrase}. The dataset includes 1,066,115 instances mapping a single complex sentence to a sequence of sentences that express the same meaning, together with RDF triples that describe their semantics. They considered two system setups: a text-to-text setup that does not use the accompanying RDF information, and a semantics-augmented setup that does. They report a BLEU score of 48.9 for their best text-to-text system, and of 78.7 for the best RDF-aware one. We focus on the text-to-text setup, which we find to be more challenging and more natural. We begin with vanilla SEQ2SEQ models with attention (Bahdanau et al., 2015) and reach an accuracy of 77.5 BLEU, substantially outperforming the text-to-text baseline of Narayan et al. (2017) and approaching their best RDF-aware method. However, manual inspection reveal many cases of unwanted behaviors in the resulting outputs: (1) many resulting sentences are unsupported by the input: they contain correct facts about relevant entities, but these facts were not mentioned in the input sentence; (2) some facts are repeated the same fact is mentioned in multiple output sentences; and (3) some facts are missing mentioned in the input but omitted in the output. The model learned to memorize entity-fact pairs instead of learning to split and rephrase. Indeed, feeding the model with examples containing entities alone without any facts about them causes it to output perfectly phrased but unsupported facts (Table 3). Digging further, we find that 99\% of the simple sentences (more than 89\% of the unique ones) in the validation and test sets \hlc[paleplum]{also appear in the training set, which coupled with the good memorization capabilities of SEQ2SEQ models and the relatively small number of distinct simple sentences helps to explain the high BLEU score}. \hlc[carrotorange]{To aid further research on the task, we propose a more challenging split of the data}. We also establish a stronger baseline by extending the SEQ2SEQ approach with a copy mechanism, which was shown to be helpful in similar tasks (Gu et al., 2016; Merity et al., 2017; See et al., 2017). On the original split, our models outperform the best baseline of Narayan et al. (2017) by up to 8.68 BLEU, without using the RDF triples. On the new split, the vanilla SEQ2SEQ models break completely, while the copy-augmented models perform better. In parallel to our work, an updated version of the dataset was released (v1.0), which is larger and features a train/test split protocol which is similar to our proposal. We report results on this dataset as well. The code and data to reproduce our results are available on Github.1 We encourage future work on the split-and-rephrase task to use our new data split or the v1.0 split instead of the original one. \\
\hline
\textbf{Talk Transcript:} Let's begin with the motivation so \hlc[dandelion]{processing long complex sentences is a hard task this is true for arguments like children people with reading disabilities second language learners but this is also true for sentence level and NLP systems} for example previous work show that \hlc[dandelion]{dependency parsers} degrade performance when they're introduced with longer and longer sentences in a \hlc[dandelion]{similar result was shown for neural machine translation} where neural machine translation systems introduced with longer sentences starting degrading performance \hlc[mediumspringbud]{the question rising here is can we automatically break a complex sentence into several simple ones while preserving the meaning or the semantics and this can be a useful component in NLP pipelines}. For example \hlc[palecerulean]{the split and rephrase task was introduced in the last EMNLP by Narayan Gardent and Shimarina where they introduced a dataset an evaluation method and baseline models for this task. The task definition can be taking a complex sentence and breaking it into several simple ones with the same meaning}. For example if you take the sentence Alan being joined NASA in nineteen sixty three where he became a member of the Apollo twelve mission along with Alfa Worden and his back a pilot and they've just got its commander who would like to break the sentence into four sentences which can go as Alan bean serves as a crew member of Apolo twelve Alfa Worden was the back pilot will close it was commanded by David Scott now be was selected by NASA in nineteen sixty three we can see that the task requires first identifying independence semantics units in the source sentence and then rephrasing those units into a single sentences on the target site. In this work we first show \hlc[paleplum]{the simple neural models seem to perform very well on the original benchmark but this is only due to memorization of the training set} \hlc[carrotorange]{we propose a more challenging data split for the task} to discourage this memorization and we perform automatic evaluation in error analysis on the new benchmark showing that the task is still very far from being solved.
\\ 
 \hline
\end{tabular}}
  \caption{Alignment example between a paper's Introduction section and first 2:40 minutes of the talk's transcript. The different colors show corresponding content between the transcript to the written paper. This is the full-text version of the example shown in Table~\ref{fig:example}.}~\label{fig:example_gt_alignment}
\end{table}

\begin{table}[t]
\resizebox{1.0\columnwidth}{!}{
\scriptsize

\begin{tabular}{ |p{2.53cm}|p{2.53cm}|p{2.53cm}|p{0.15cm}| } 
\hline
\textbf{Paper Sentence} & 
\textbf{ASR transcript} &
\textbf{Human transcript} &

\\ 
\hline
Processing long, complex sentences is challenging. & 
base begin motivation processing long complex sentences hard task &
Let's begin with the motivation so processing long complex sentences is a hard task & 
\greenv
\\ \hline
This is true either for humans in various circumstances or in NLP tasks like parsing and machine translation. & 
true arguments like children people reading disabilities second language learners also true first sentence level p system &
this is true for arguments like children people with reading disabilities second language learners but this is also true for sentence level and NLP systems & 
\greenv
\\ \hline
A recent work by Narayan et al. (2017) introduced a dataset, evaluation method and baseline systems for the task, naming it Split-and Rephrase. &
previous work show data tendency parsers great performance introduced longer longer sentences &
previous work show that dependency parsers degrade performance when they're introduced with longer and longer sentences &
\redx
\\ \hline
This is true either for humans in various circumstances or in NLP tasks like parsing and machine translation. &
similar results showing new machine translation new machine translation &
similar result was shown for neural machine translation where neural machine translation &
\greenv
\\ \hline
An automatic system capable of breaking a complex sentence into several simple sentences that convey the same meaning is very appealing. &
systems introduced longer sentences starting performance question rising automatically break complex sentence several simple ones preserving meaning semantics useful company p like example &
systems introduced with longer sentences starting degrading performance the question rising here is can we automatically break a complex sentence into several simple ones while preserving the meaning or the semantics and this can be a useful component in NLP pipelines for example &
\greenv
\\ \hline
A recent work by Narayan et al. (2017) introduced a dataset, evaluation method and baseline systems for the task, naming it Split-and Rephrase. &
leader task introduced last 'll bynari guard going marina introduced data sets evaluation method baseline models task &
the split and rephrase task was introduced in the last EMNLP by Narayan Gardent and Shimarina where they introduced a dataset an evaluation method and baseline models for this task &
\greenv
\\ \hline
An automatic system capable of breaking a complex sentence into several simple sentences that convey the same meaning is very appealing. &
phoenician taking complex sentences break several simple ones example take sentence alan joined nasa nineteen sixty three became member apollo twelve mission along word inspect pilot got commander would like break sentence sentences go alan serves crew member twelve word better polls commanded david scott selected nasa nineteen sixty three &
the task definition can be taking a complex sentence and break it into several simple ones for example if you take the sentence Alan being joined NASA in nineteen sixty three where he became a member of the Apollo twelve mission along with Alfa Worden and his back a pilot and they've just got its commander who would like to break the sentence into four sentences which can go as Alan bean serves as a crew member of Apolo twelve Alfa Worden was the back pilot will close it was commanded by David Scott now be was selected by NASA in nineteen sixty three &
\\ \hline
A recent work by Narayan et al. (2017) introduced a dataset, evaluation method and baseline systems for the task, naming it Split-and Rephrase. & 
see task requires first identifying independence imagic units &
we can see that the task requires first identifying independence semantics units &
\\ \hline
The dataset includes 1,066,115 instances mapping a single complex sentence to a sequence of sentences that express the same meaning, together with RDF triples that describe their semantics. & 
source sentence rephrasing units single sentences target &
in the source sentence and then rephrasing those units into a single sentences on the target site &
\\ \hline
Digging further, we find that 99\% of the simple sentences (more than 89\% of the unique ones) in the validation and test sets also appear in the training set, which coupled with the good memorization capabilities of SEQ2SEQ models and the relatively small number of distinct simple sentences helps to explain the high BLEU score. &
work first show simple neural models seem perform well original benchmark due memorization training set &
In this work we first show the simple neural models seem to perform very well on the original benchmark but this is only due to memorization of the training set & 
\greenv
\\ \hline
To aid further research on the task, we propose a more challenging split of the data. &
perform close challenging data split task discourage instant memorization perform automatic evaluation analysis new benchmark showing task still far &
we propose a more challenging data split for the task to discourage this memorization and we perform automatic evaluation in error analysis on the new benchmark showing that the task is still very far from being solved &
\greenv
\\ \hline
\end{tabular}}
  \caption{Alignment obtained using the HMM, for the Introduction section and first 2:40 minutes of the video's transcript.}~\label{fig:example_model_output}

\end{table}

\end{document}